# Representation learning for a generalized, quantitative comparison of complex model outputs

Colin G. Cess[1] and Stacey D. Finley[1,2,3]*

[1]Alfred E. Mann Department of Biomedical Engineering, University of Southern California, Los Angeles, CA
[2]Department of Quantitative and Computational Biology, University of Southern California, Los Angeles, CA
[3]Mork Family Department of Chemical Engineering and Materials Science, University of Southern California, Los Angeles, CA

**ABSTRACT**

Computational models are quantitative representations of systems. By analyzing and comparing the outputs of such models, it is possible to gain a better understanding of the system itself. Though as the complexity of model outputs increases, it becomes increasingly difficult to compare simulations to each other. While it is straightforward to only compare a few specific model outputs across multiple simulations, additional useful information can come from comparing model simulations as a whole. However, it is difficult to holistically compare model simulations in an unbiased manner. To address these limitations, we use representation learning to transform model simulations into low-dimensional points, with the neural networks capturing the relationships between the model outputs without the need to manually specify which outputs to focus on. The distance in low-dimensional space acts as a comparison metric, reducing the difference between simulations to a single value. We provide an approach to training neural networks on model simulations and display how the trained networks can then be used to provide a holistic comparison of model outputs. This approach can be applied to a wide range of model types, providing a quantitative method of analyzing the complex outputs of computational models.

## 1 INTRODUCTION

As the modern century marches forward, computing power continues to increase. At the same time, advances in experimental techniques enable one to capture enormous amounts of data[1–5]. By taking advantage of these powerful computing resources and large experimental datasets, scientists can construct larger and more complex computational models of various systems[6–13]. These models allow us to interrogate a system in extensive detail, predicting how the system responds to various perturbations. In some cases, models enable us to examine pieces of the system that cannot yet be experimentally measured. In addition, models permit us to perform large-scale simulations to produce *in silico* data that would either take too much time or be too costly to generate experimentally.

To understand the system being modeled, these model simulations must be properly analyzed. Many methods currently exist for analyzing computational models and *in silico* data. Some methods, such as sensitivity analyses or parameter sweeps, are designed specifically to examine a model[14–21]. Other methods, such as partial least squares and clustering, are typically used to analyze data[22–25] but can easily be used to analyze models as well[26]. However, a current drawback of most model analysis methods is that they generally examine only a piece of the system, rather than the whole set of simulated data[27–31]. For example, performing a sensitivity analysis requires one to explicitly define the output of interest, and the output selected influences the analysis itself. Consider a system that oscillates with time. For one state variable, we can easily calculate outputs such as the amplitude and frequency of peaks. But as the number of state variables increases, to get a complete view of the system, we would have to account for how the amplitudes and frequencies for each variable relate to those of every other variable. To be complete, we must also consider the possibility that a state variable may not oscillate at all given the parameter values being simulated. Additionally, we would have to scale the different comparisons to eliminate potentially unjust influence based on output values. This could introduce bias from the researcher, who would have to manually determine these comparison metrics and could easily fail to account for every possible comparison. This example highlights limitations in obtaining a holistic and unbiased comparison of model outputs.

To address these limitations, in this study, we present a generalized, model-agnostic approach for comparing complex model outputs as a single, continuous value by comparing learned representations of the outputs. Representation learning, in general, uses neural networks to project inputs, such as images and text, into lower-dimensional feature-vectors, which can then be used for various downstream tasks[32]. One specific use of representation learning is Siamese neural networks. Siamese networks are a pair of identical neural networks that project a pair of inputs into individual, low-dimensional points. By taking the distance between these points, the similarity between the inputs can be determined[33,34]. This implementation has been used to detect signature forgeries[35], to visualize single-cell data[36], for face verification[37], and as a measure of continuous disease progression[38]. It has also been also been shown to work very well for dimensionality reduction[39] and for clustering mass spectra[40].

Here, we use representation learning with a Siamese implementation to compare the outputs of complex computational models. By projecting outputs to low-dimensional space and taking the distance between points, we can determine how different two simulations are. The distance between the points provides a single, holistic value that accounts for the complex relationships between model output features that would be otherwise difficult to calculate manually. Our approach is unique in that we are applying a method traditionally used to analyze real-world data to analyze computational models. Other works have applied deep-learning methods to model-generated data; however, that was to better analyze experimental data, not to analyze a model itself[41]. Some methods exist that compare simulated time series data[42,43]; however, these are not applicable to other types of model outputs. We display our approach on several example models that have very different types of outputs to show its broad applicability.

## 2 RESULTS

We first clarify some terminology to eliminate potential confusion throughout the paper. The term "model" is used to refer to any sort of computational model, such as systems of differential equations, that can produce a "model output", any simulation or calculation performed by a model. These two terms, model and model output, specifically refer to the computational models that are being analyzed, and not to the neural networks that are used to perform the analysis. The terms "projected space" and "projections" is used to refer to the learned representations of model outputs that are used to perform analyses.

Here, we describe our approach and provide three example models that produce very different types of outputs. Our goal with this study is to display the utility of our approach, potential ways that it can be used to analyze a variety of models, and areas where it could be improved. We briefly discuss each model and the analysis results so that the implementation of the approach can be understood in the context of the model. However, we avoid discussing the implications of the results on the modeled systems, as our aim is to demonstrate that this approach can be applied to disparate models and not to draw novel insights about the modeled systems.

## 2.1 Representation learning: Overview and training

Our comparison approach makes use of neural networks to reduce the dimensionality of model outputs and then calculate the distance between the projected points to determine how different they are from each other[33,34]. A schematic of this is displayed in **Figure 1**. There exist many different ways to train a neural network for representation learning. We use a modified version of SimCLR[44]. SimCLR uses an encoder to convert inputs into representations, followed by a projection head, with loss calculated on the projections. After training, the projection head is removed, and the representations are used to train a linear classifier. The key aspect of training is that, in each training batch, each input is augmented twice, sent through the network. The projected augmentations are compared to each other and the augmentations from all other inputs using normalized temperature-scaled cross entropy loss (NT-Xent) with cosine similarity. As expanded upon in the Methods, we make two changes from this approach. Because our downstream task is the use of the distance between representations, we remove the projection head and calculate loss directly on the representations. We also replace cosine similarity with 1/(1 + Euclidean distance), as Euclidean distance is our main measurement.

In practice, neural networks are often used in ensembles. For the results presented below, we report the mean of the distances projected by an ensemble of 50 neural networks, along with their standard deviations.

## 2.2 Reconstruction of 2-dimensional data

Before applying our method to high-dimensional, model-generated data, we first perform a simple test to confirm that our proposed method works as intended. Following Szubert et al[36], we generate two sets of two-dimensional data (**Figure 2, top row**) (*x, y*), which are then transformed into nine-dimensional space (*x+y, x-y, xy, $x^2$, $y^2$, $x^2y$, $xy^2$, $x^3$, $y^3$*). We then use our approach to learn two-dimensional representations of this transformed dataset, aiming to reconstruct the original structure (**Figure 2, middle row**). We find that our approach performs a reasonable reconstruction of the original data, meaning that the learned representations have useful information.

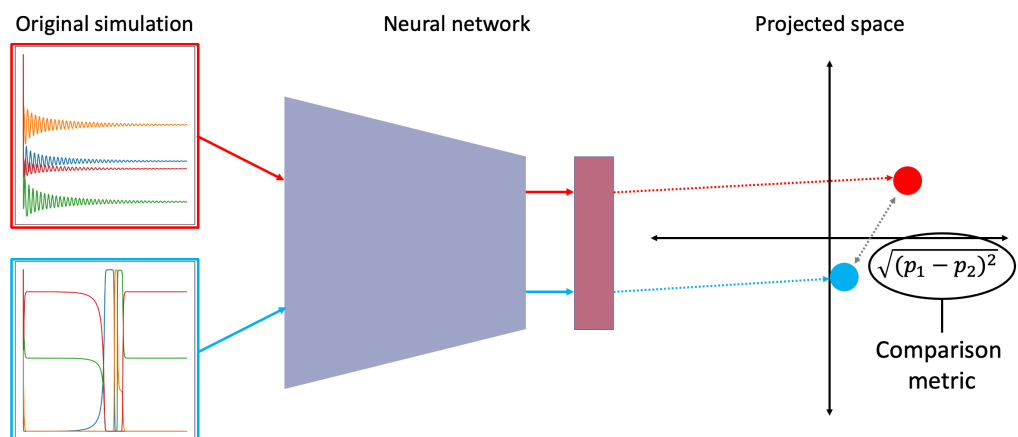

**Figure 1.** Schematic displaying how two simulations are compared via projection to learned space.

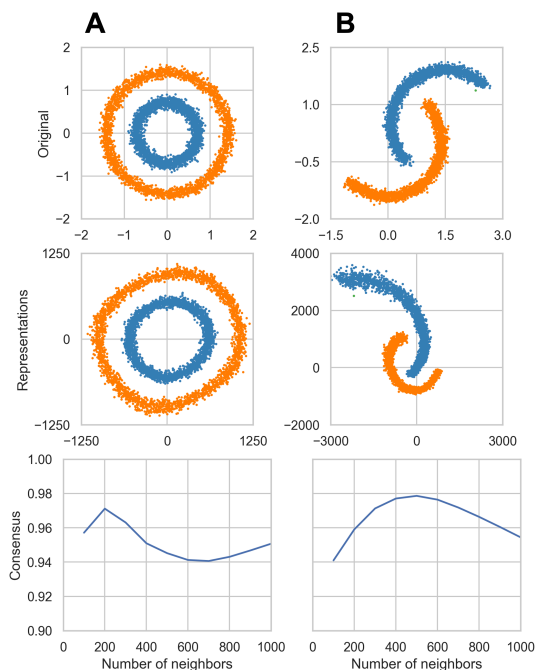

**Figure 2.** Test dataset A and B. Top row: original data. Middle row: learned representations from 9-d transformation of original data. Bottom row: consensus scores for a range of neighborhood sizes.

## 2.3 Consensus score

Because neural networks are stochastically initialized and trained, different networks can potentially yield different representations. Additionally, unlike the test set in **Figure 2**, we do not know beforehand what the learned representations should look like. Therefore, we developed the "consensus score" metric to evaluate how different the learned representations for multiple neural networks are from each other. The higher the consensus score, the more similar the learned representations. As the analyses that we wish to perform are based on the distance between points in learned space, a higher consensus between networks is favorable, as it means that there is higher agreement between the networks in an ensemble, increasing confidence in the results.

Our consensus score (expanded upon in the methods) compares the local spatial layouts of two sets of projections. It is calculated using the fraction of $n$ nearest neighbors to a point that are shared for projections from two networks. For example, if the three nearest neighbors of point $p$ are points [2, 3, 5] for network A, and points [2, 3, 4] for network B, then the consensus for that point for those two networks is 0.66. We perform this calculation for each point and average the results to get the overall consensus between two networks. We then average all of the pairwise consensus scores between each network in an ensemble to get the overall consensus for that neural network structure for a particular set of simulations. The consensus scores for both of the above test datasets at several values of $n$ are shown in **Figure 2, bottom row**. In general we would expect the consensus score to increase as we increase $n$. However, the consensus for these two examples already display very high scores and are varying only over a range of 0.04. Thus, we believe we have developed a robust metric for characterizing the similarity between results from distinct neural networks.

## 2.4 Overview of test models

We test our approach on three disparate computational models: a constraint-based metabolic flux model, an ordinary differential equation (ODE) model, and a spatial agent-based model (ABM). These models (detailed in the following sections) were chosen because they produce outputs in different formats, which are, respectively: a vector of fluxes for each reaction (1D vector), time-series for each ODE species (2D matrix), and a spatial layout for each agent type (3D matrix). Because of the different output formats, we use different types of neural networks for each model.

For analyzing each model, we use ensembles of 50 neural networks, each trained on 10,000 simulations. For all three test models, we project to 16 dimensions, with the final layer of neurons having a linear activation. The consensus for each model is shown in **Figure 3**. As expected, the consensus was lower than it was for the datasets used in **Figure 2** (as model outputs were not originally in a low dimension) and increases as the number of neighbors used to calculate score increases. Interestingly, there was a lower consensus between the neural networks used to represent the flux model, compared to the other two models. This result was unexpected, as this model has the simplest output format. Further work is needed to explore this specific model type and find methods that can produce greater consistency between representations. However, as our primary goal is to demonstrate the use of our method to represent complex model outputs to enable standard analyses, we present example analyses that may be performed using the results from our approach.

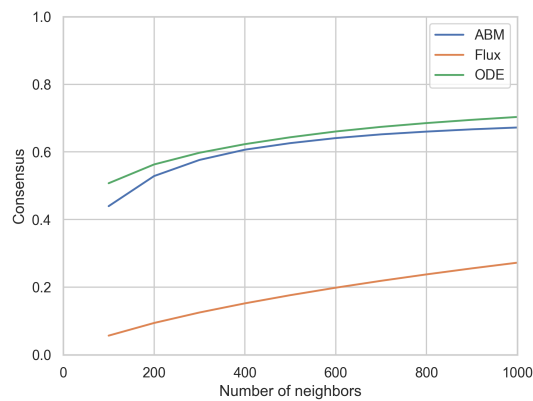

**Figure 3.** Consensus scores for the three test models.

In the following three sections, we detail each model and show example analyses that can be performed using this approach to better understand model behavior. We display two analyses for each model. The first is the same for each, and that is shifting the value of one parameter to see how the distance from the base parameter set changes. We use this same analysis to show how this approach translates to different model types. The second test is different for each model and is one that could be of interest to that specific model type. However, these specific tests can also be used for other model types.

## 2.5 Example 1: Flux model

The first example uses a model of metabolic reactions for *E. coli*, containing 2,477 reactions organized into 44 subsystems[68]. Here, the model parameters are the upper and lower flux bounds for each reaction. These bounds are fixed, and the model estimates the reaction fluxes within the bounds needed to optimize a specified objective function[69]. Positive fluxes represent flow in the forward direction, while negative fluxes are flow in the reverse direction. The output for this model is a one-dimensional vector of the flux through each reaction for the optimized state. Because the output is a simple vector, we use a basic feedforward neural network to analyze this model.

For the first test, we shift the lower bound for uptake of uridine diphosphate glucose (a nutrient for the organism) from -1,000 to 0 (in nutrient uptake, negative flux corresponds to flow into the cell) and compute the distance from the optimized flux state to that of when the bound is 0. This allows us to see how the overall metabolic state changes as the cell becomes able to take up more of this nutrient. We find that there is a steep increase in the distance from the base state once the organism becomes able to uptake this nutrient (**Figure 4**), meaning that this nutrient causes a distinct change in the metabolic pathways utilized, compared to the base state. This difference continues to increase as *E. coli* can take up more of the nutrient, until it finally levels off, with increased potential uptake no longer changing the metabolic state.

The second test we perform, which is very specific to this type of model, is a series of reaction knockouts[70–72]. Here, we set the upper and lower bounds for the flux through a specific reaction equal to zero before solving the model for fluxes through the rest of the reactions (**Figure 5**). This means the organism cannot utilize that particular reaction, and it has to direct flux through alternative pathways. Because of the large number of reactions in this model, it would be difficult to compare the effect of each individual knockout on the estimated flux through each metabolic reaction. Instead, we examine the average change for knockouts in each subsystem. For each subsystem, we knockout each reaction individually, compute the distance to the base state, and then average the distances for each reaction in the subsystem. This allows us to determine which subsystems contribute the most to the base state and thus could be subject to further analysis.

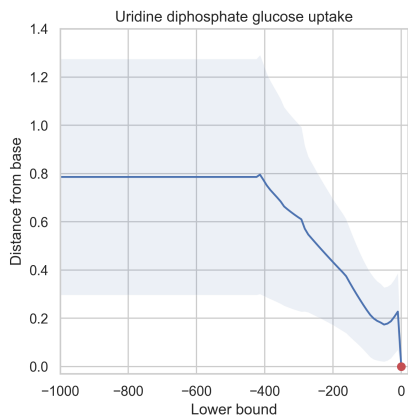

**Figure 4.** Change in model state (as distance from the simulation produced at a lower bound of zero, calculated by a Siamese network) when changing the lower bound of a nutrient uptake. Line – mean; shaded area – standard deviation.

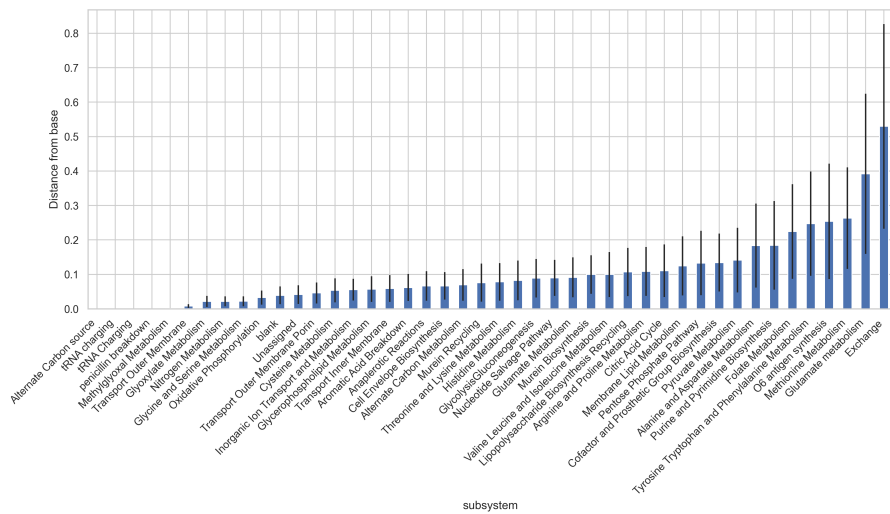

**Figure 5.** Knockout results from the flux model, averaging the distance from the base state for each subsystem. Error bars show standard deviation.

## 2.6 Example 2: Ordinary differential equation model

The second example is a Lotka-Volterra model, which is comprised of ordinary differential equations. This model has been used in many fields, such as ecology, chemistry, and economics[45–47]. We chose this model because of its broad usage and because, depending on parameter values, it can reach a steady state or produce oscillations. Specifically, we use a four-species model, setting the base parameters to a set listed in Vano et al., who examined chaotic behavior in this system[48]. We increase three of the parameter values from zero to 0.01 (shown in equation (3) of Vano et al.), which maintains the oscillatory behavior, but allows us to sample these parameters above and below the base value of 0.01 when producing the training dataset. The output of this model is a vector of values at each time point for each species. We organize the model outputs as a two-dimensional matrix, $timepoints \times species$. The neural network we use for this type of model is a one-dimensional convolutional neural network, which has been used to classify time-series data[49,50].

The first test that we perform is shifting the value of a parameter; here, from two-fold below to two-fold above the base value. We calculate the distance between the output from each parameter value and the base output (**Figure 6**). As we would expect, the further the parameter value is shifted from the base value, the larger the predicted distance is. While this is a simple test, it shows how we can use this approach to easy perform a holistic evaluation of how model behavior changes as a parameter value is changed.

Our second test for this model is a local sensitivity analysis (**Figure 7**), which is used to determine how sensitive a model output is to changes in a parameter. A local sensitivity analysis changes one parameter at a time from a base value (here, increasing by 10%) and records the corresponding change in a specified output. We consider two cases: (1) using the distance in projected space to holistically compare the model outputs and (2) specifying a single model output. In projected space, we calculate the distance between the output from the perturbed parameter and the base output to get an overall comparison between the two. For the specified output, we use the average change of all four final values of the differential equations in the model. This output is only capturing one moment in time, and fails to account for temporal behaviors, such as the oscillations that this model is able to produce. One common usage for a sensitivity analysis is to determine which parameters most strongly impact the output, and thus are most important for parameter estimation. With this in mind, we averaged the sensitivity results for increasing each parameter value for each case of sensitivity analysis (representations and specified output) and ranked the parameters by the most sensitive. Because of the difference in how we compare perturbed simulations to the baseline, we normalize the sensitivities to between 0 and 1 so that we can compare them qualitatively. Comparing the two cases, we find that some parameters (3, 16) have similar rankings, whereas other parameters have very different rankings. Specifically, we draw attention to parameters 6 and 19. In **Figure 7**, the box in the top right shows the base simulation, while the other insets show the simulations for a 10% increase in parameters 6 (left) and 19 (right). The specified output ranks parameter 19 as slightly more influential while using learned representations ranks parameter 6 as having a much higher influence on the model output. From a visual comparison to the base simulation, one would likely argue that perturbing parameter 6 yields a larger overall change in the model output than parameter 19. With this, we show how our approach allows us to get a better idea of how perturbing a parameter impacts model

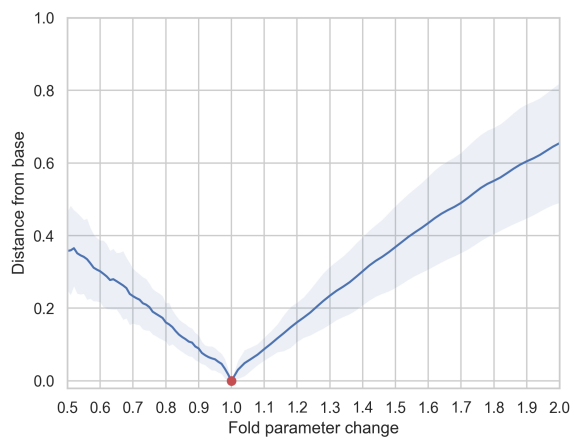

**Figure 6.** Shifting the value of a single parameter over a two-fold range (x-axis) and calculating the distance from the base model output (red point) parameter (y-axis). Line – mean; shaded area – standard deviation.

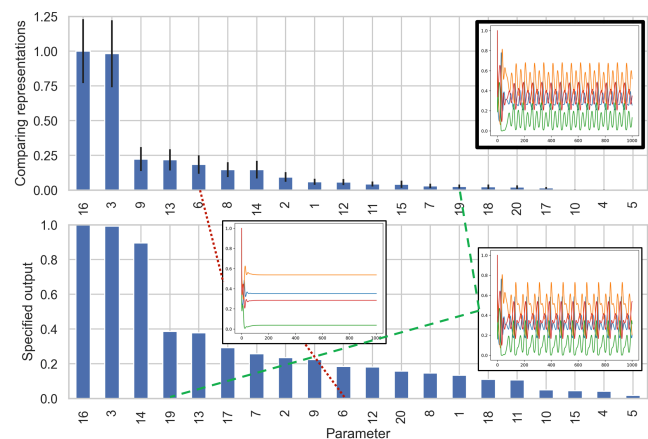

**Figure 7.** Local sensitivity analysis for the second test model. Compares the most sensitive parameters when using a Siamese network as the output (top) to using the final value of each variable as the output (bottom). Values are normalized to the maximum for each sensitivity output. Error bars (top) show standard deviation. Inset in the top right shows the base simulation. The other insets show model simulations following a 10% increase for parameter 6 (left) and parameter 19 (right).

behavior as a whole, instead of looking at a small region of parameter space.

## 2.7 Example 3: Spatial agent-based model

The final example is an agent-based model (ABM) of tumor-immune interactions[51]. ABMs are used in many diverse fields to explore how interactions between individuals yield complex, emergent behaviors at the population level[52–62]. This ABM is an on-lattice model consisting of three cell types that can take on a total of six states. The output is the final spatial layout, which we format as a three-dimensional matrix. The first two dimensions are the lattice dimensions of the model, and the third dimension is for each cell state. Because this is a similar format to an image, with the number of cell states being akin to color-channels, we used a two-dimensional convolutional neural network[63–66].

As with the previous example, we first performed a single parameter shift. We varied the tumor cell proliferation rate across 21 different values (**Figure 8**), with the base simulation being the middle value. Because this is a stochastic model, we performed each simulation with ten replicates. When taking the distance between simulations for two different parameter sets, we compared each replicate to every other replicate and then took the average distance between them. We again find that the distance between model outputs increases as we shift the parameter value.

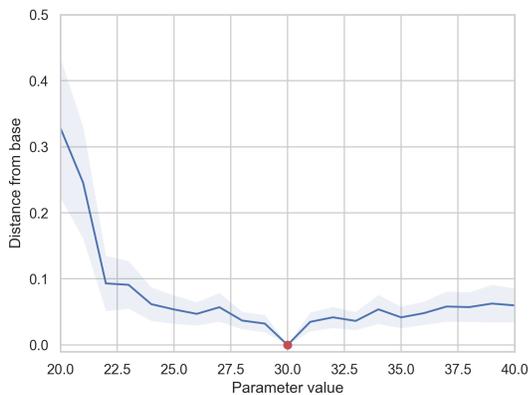

**Figure 8.** Shifting the value of one parameter (x-axis), comparing the distance calculated by a Siamese network to the middle parameter value (red point on the y-axis). Line – mean; shaded area – standard deviation

The second test we performed was clustering the model Monte Carlo simulations that produced the training dataset (**Figure 9**). We use the distances in projected space to group the 10,000 simulations via hierarchical clustering[67]. From these clusters, we examined the distributions for model parameters that produced the simulations (top two rows). We looked at different numbers of clusters, with two producing clear separation in the parameter distributions. We see clear differences in the distributions for the parameters, allowing us to identify how different model end-states compare to each other based on model parameter values. Additionally, we can then look at selected specific model outputs (bottom row). This gives us insight into some of the systemic characteristics of each cluster. For example, we see that, in the orange cluster, low numbers of cancer cells are clustered with high numbers of T cells, which kill cancer cells. Furthermore, these two outputs are clustered with a low macrophage recruitment rate, which makes sense as, in this model, macrophages suppress T cell activity. This example demonstrates that our approach can be used to cluster model simulations and examine how overall model output is linked to model parameters and specific outputs.

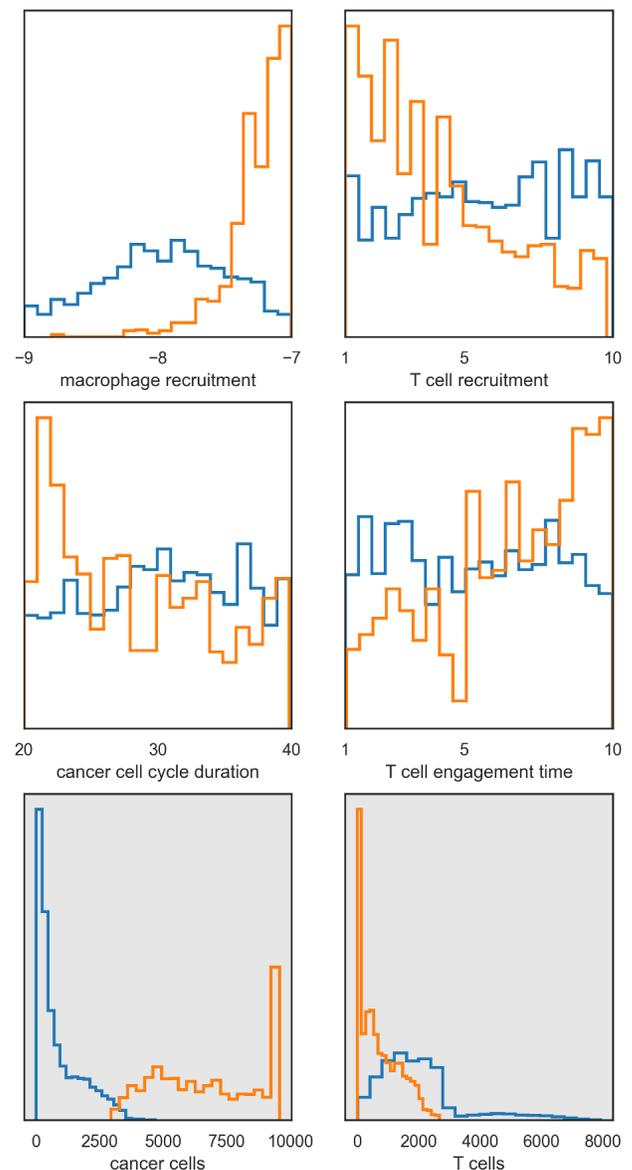

**Figure 9.** Clustering Monte Carlo simulations for the third test model. Distributions for four sampled parameters are shown for each of the two clusters in the white plots (top two rows) and for two selected model outputs in the gray plots (bottom row).

## 3 DISCUSSION

With this study, we display how representation learning can be used to compare complex model simulations as a single quantitative value. We show how the same method can be applied to three very different types of models, only needing to change the neural network structure to accommodate the format of the model output. By using a neural network to reduce the model output to a low-dimensional point and taking the distance between two points, we can obtain a holistic view of how different two simulations are from each other, without the need to manually calculate complex comparison metrics.

To display the utility of our approach, we provided a total of six tests performed across three example models, showing how the approach allows us to compare model simulations in an unbiased manner. The first test for each model was the same, where we shift the value of a single parameter across a range of values. This simple test displays how model outputs change overall as a parameter is varied. This analysis can be used to determine areas of parameter space where the model output changes more drastically, or the identify potential regions where biphasic behavior is exhibited. For the first example model, we also displayed how our approach can be applied to reaction knockout simulations. Often, these simulations are used to identify perturbations that optimize a specific biological process[70–72]. Our approach can extend these analyses to determine how different the new metabolic states are from the original or from other knockout states. Our second example model is a time series differential equation model used in a variety of fields. We show how our approach can be used to aid in sensitivity analyses[18], which can then identify important parameters for parameter fitting, model reduction, and model expansion. Lastly, we used our approach to cluster simulations from an agent-based model. From these clusters, we were able to examine how different parameter values tend to create different simulation end-states, along with how specific model outputs are linked to the different end-states. In the context of the tumor model that we examined, our approach would allow the researcher to better predict how different tumor properties yield different final tumor states.

We only displayed simple examples of analyses so that our approach could be easily understood; however, it can be extended to more complex analyses. For example, the sensitivity analysis that we showed was a simple, local analysis. However, this only captures a small region of parameter space, whereas global sensitivity analyses yield a better exploration of the model[21,73,74]. Instead of comparing to a base parameter set, each dimension of the projected points can be treated as an output for a global sensitivity analysis, providing a more exhaustive description of how sensitive the model as a whole is to each parameter. This approach could also be used to aid in uncertainty and robustness analyses[75–78]. By accounting for model behavior as a whole instead of only focusing on a small part of the output space, our approach provides a better characteristic of the system. Another potential use is for optimizing a perturbation to a system that would produce a desired change in an output of interest while minimizing the overall change in the system.

We note that we did not explore the many other ways to perform representation learning. Potentially other methods can improve the consistency of the learned representations, which would improve confidence in the calculated distance between representations. For example, one could consider implementing different training methods, variational autoencoders, or supervised pre-training, where we would first predict model parameters from the model outputs[79–84]. The purpose of this work, however, is to present a novel application of representation learning, and we leave extensive analyses to future work, where our approach can be optimized in more case-specific settings.

The main limitation with our approach is that it can be computationally expensive. One can quickly generate large numbers of training samples for small models; however, as model complexity increases, the time required to perform Monte Carlo simulations increases greatly. Additionally, it takes time to determine an optimal neural network structure, especially when more complex neural networks need to be implemented, based on the format of the model outputs. However, once a neural network is trained, it can be used for many different analyses. Another limitation is the nature of neural networks. Because they are black-box methods, it is difficult to interpret how model outputs are being projected into low-dimensional space. Our approach simply determines how different two simulations are, not why they are different. However, as we show with the third example, combining this approach with specific model outputs helps yield further insight into the system.

Despite these limitations, we present a useful approach to compare model outputs via representation learning. This work addresses the limitations of commonly used model analysis methods; namely, a narrowly focused and potentially biased comparison. We show how these neural networks can be trained on a set of model-generated data and how the trained networks can be applied to analyze a range of model outputs. We demonstrate that this approach provides an additional way to interrogate models beyond current methods.

## 4 METHODS

The approach we detail here uses representation learning to train neural networks to project the outputs of computational models into low-dimensional space, where the neural networks can then be applied in a Siamese fashion to compare model outputs as a single, continuous value. In the following sub-sections, we describe: (1) an overview of representation learning, (2) our specific training approach, (3) generation of training data and specific neural network construction, (4) calculation of a "consensus score," and (5) implementation for analyzing model outputs.

***Overview of representation learning.*** The approach we detail here uses representation learning to train neural networks to project the outputs of computational models into low-dimensional space, where the neural networks can then be applied in a Siamese fashion to compare model outputs as a single, continuous value. In the following sub-sections, we describe: (1) an overview of representation learning, (2) our specific training approach, (3) generation of training data and specific neural network construction, (4) calculation of a "consensus score," and (5) implementation for analyzing model outputs.

***Training approach.*** Many methods exist for training neural networks for representation learning[32]. Here, we use the training method proposed in SimCLR[44]. This method aims to maximize the similarity of the projections for two augmentations of an input in comparison to their similarity to the other inputs. We chose this method because it does not require inputs to be labeled as similar or dissimilar to each other, whereas training methods such as triplet loss require explicitly labeling training samples as similar or dissimilar[85,86].

We make two changes to this approach. SimCLR, like many representation learning frameworks, makes use of a projection head that follows the learned representations, with loss being calculated based on the cosine similarity between the outputs of this projection head. The use of a projection head has been found to improve accuracy when training a linear classifier on the learned representations of images[44]. However, our aim is to instead compare distances in projected space. Therefore, like many studies using Siamese networks, we do not use a projection head and instead calculate loss directly on the learned representations. Additionally, we use Euclidean similarity instead of cosine similarity, as shown in **Equation 1**, where $d(p_1, p_2)$ is the Euclidean distance between points in projected space. The reason for this is that we are projecting to a low number of dimensions, where we are using the distance between simulations as a measure of comparison. We want this comparison to be unbounded, hence the use of Euclidean distance instead of cosine.

$$similarity = \frac{1}{1+d(p_1,p_2)} \quad (1)$$

***Generation of training data and neural network structure.*** We take a straightforward approach to generate training data. We simply perform Monte Carlo simulations with each computational model, sampling parameters from a uniform distribution across the entire range of parameter space that we may want to vary them over. This produces a wide range of model outputs for training the neural network. Each model simulation is treated as a unique individual, whether or not the model is stochastic. In this study, we perform 10,000 simulations. Sampling ranges were chosen based on the specific model being analyzed.

The main part of our approach that is model-specific is the neural network structure, which must be chosen based on the format of the model output. For temporal models, a one-dimensional convolutional neural network is often sufficient, whereas a two-dimensional convolutional network is suitable for most spatial outputs. More complex networks may be suitable based on the nature and scope of the model's output.

***Consensus score.*** One of the issues with neural networks is that they have large numbers of parameters and are stochastically initialized, meaning that each time a neural network is trained, the values of neuron weights are different. Because of this, different representations may be learned each time a network is trained. Therefore, we developed what we term the "consensus score" between two trained neural networks, which quantifies how similar the projections are to each other. We make the assumption that if multiple stochastically initialized neural networks converge to similar projections, then we can have more confidence in those representations.

This score is calculated in the following way. For each neural network in an ensemble, we project the training data into low-dimensional space. For each projected point, we find its nearest $n$ neighbors, referring to this as the point's neighborhood. To determine the consensus between neural networks A and B, for each point, we calculate the fraction of the neighborhoods that are the same based on their projections from A and B. We average this across all points. This then gives us the consensus between networks A and B.

We calculate the pairwise consensus score between each neural network in an ensemble and average them to get the overall consensus within the ensemble. We select neural network structures that maximize this score.

***Implementation for analyzing model outputs.*** By using the trained neural networks in a Siamese implementation, we can use the distance between projected points as a measure of how different they are, with closer points being more similar to each other than they are to distant points. The main advantage of this approach is that, by using a neural network to learn representations of model outputs, there is no need to manually calculate a difference metric. We note that the distance between two projected points can be influenced by the neural network structure and training, and thus has little meaning by itself. Instead, comparisons between multiple outputs are needed to give it meaning. For example, for outputs $O_1$, $O_2$, and $O_3$, the distance between $O_1$ and $O_2$ alone provides little information, however if that distance is smaller than the distance between $O_1$ and $O_3$, we conclude that $O_1$ is more similar to $O_2$ than it is to $O_3$.

For all three test models, we project to 16 dimensions. This final layer of neurons uses a linear activation, as we are performing analyses directly on this layer. Other neural network hyperparameters were adjusted to maximize the consensus score.

## ACKNOWLEDGEMENTS

The authors thank members of the Finley research group for critical comments and suggestions.